\documentclass{edm_article}
\usepackage{color,soul}
\usepackage{makecell} 
\usepackage{titlesec}
\usepackage{hyperref}
\usepackage{listings}
\usepackage{xcolor}
\usepackage{geometry}
\usepackage{placeins}
\usepackage{verbatim}
\usepackage{lipsum} 
\usepackage[numbers]{natbib}
\usepackage{fancyhdr} 

\newcommand{\footnotetextcolor}{\color{blue}}

\begin{document}


\title{Combining Cognitive and Generative AI for Self-explanation in Interactive AI Agents}
\author{
  Shalini Sushri \\
  \affaddr{Georgia Institute of Technology}\\
  \email{ssushri3@gatech.edu}
  \and
  Rahul K. Dass\\
  \affaddr{Georgia Institute of Technology}\\
  \email{rdass7@gatech.edu}
  \and
  Rhea Basappa\\
  \affaddr{Georgia Institute of Technology}\\
  \email{rb324@gatech.edu}
  \and
  Hong Lu\thanks{Work done as a Research Scientist at Georgia Institute of Technology.}\\
  \affaddr{Tufts University}\\
  \email{hlu07@tufts.edu}
  \and
  Ashok K. Goel\\
  \affaddr{Georgia Institute of Technology}\\
  \email{ashok.goel@cc.gatech.edu}
}
\maketitle

\vfill 

\begin{flushright}
  \footnotesize
  Appears on the Website of Human-Centric eXplainable AI in Education (HEXED) Workshop held in conjunction
  with the Seventeenth International Conference on Educational Data Mining (EDM), July 2024.
\end{flushright}
\

\begin{sloppypar}
\begin{abstract}
The Virtual Experimental Research Assistant (VERA) is an inquiry-based learning environment that empowers a learner to build conceptual models of complex ecological systems and experiment with agent-based simulations of the models.
This study investigates the convergence of cognitive AI and generative AI for self-explanation in interactive AI agents such as VERA.
From a cognitive AI viewpoint, we endow VERA with a functional model of its own design, knowledge, and reasoning represented in the Task--Method--Knowledge (TMK) language. 
From the perspective of generative AI, we use ChatGPT, LangChain, and Chain-of-Thought to answer user questions based on the VERA TMK model.
Thus, we combine cognitive and generative AI to generate explanations about how VERA works and produces its answers.
The preliminary evaluation of the generation of explanations in VERA on a bank of 66 questions derived from earlier work appears promising. 
\end{abstract}

\keywords{Self-explanation, AI Agents, Combining Cognitive and Generative AI, Theory of Mind} 

\section{Introduction}

\subsection{self-explanation in Interactive AI Agents}
Interactive AI agents with self-explanation capabilities foster understanding, transparency, and trust in users across a wide range of domains and applications \cite{lombrozo2006structure,muller2021}.
By self-explanation, we mean Interactive AI agents that can explain their reasoning and behaviors.
By generating human-understandable explanations, self-explainable AI can enhance user learning and trust \cite{elton2020}. Studies have shown the benefits of self-explanation in multimedia learning environments, facilitating intrinsic motivation, visual processing, and learning outcomes \cite{wang2023}. Additionally, emerging methods leveraging situation awareness holds promise for generating explanations of autonomous agents' behaviors, ultimately improving trust and comprehension \cite{dazeley2021}.

 This research contributes to the goal of enhancing user trust and learning through self-explanation in the Virtual Experimental Research Assistant (VERA; \cite{an2018vera, an2020vera}), an interactive learning environment for inquiry-based learning. In this paper, we explore how VERA explains its internal workings to users, potentially fostering trust and enhancing the learning experience. 

\subsection{VERA: Inquiry-based Modeling}

VERA (\url{http://vera.cc.gatech.edu}) is an interactive learning environment for supporting inquiry-based learning.
It helps learners construct conceptual models of ecological systems and evaluate them through agent-based simulations.
VERA is an AI agent because of three capabilities.
First, it uses an ontology of the ecology domain in the representation and construction of conceptual models.
Second, it automates the retrieval of species' and related ecological relations' information from the Smithsonian Institute's Encyclopedia of Life (EOL; eol.org), a comprehensive digital library of biodiversity \cite{parr2014encyclopedia}, and automatically inserts parameter values for the agent-based simulations.
Third, it automatically compiles the conceptual model into agent-based simulations in NetLogo \cite{tisue2004netlogo}.
Thus, this platform aligns with the research focus on self-explanation in educational AI assistants.

\subsection{Cognitive and Generative AI Convergence}
This research explores the potential of combining Cognitive AI and Generative AI approaches for self-explanation capabilities in VERA.
Cognitive AI is centered around understanding human cognitive processes and developing cognitively-inspired AI agents, while Generative AI methods demonstrate powerful capabilities for various natural language processing tasks like entity recognition, intent classification, and question-answering based on a text corpus \cite{liu2023combining}.


\section{Related Work}
Early research on self-explanation in Interactive AI agents highlighted the importance of explicitly representing the agent’s knowledge of its design \cite{chandrasekaran1989explaining,goel2011meta}. 
This explicit representation allows the generation of explanations about the tasks the agent performs, the domain knowledge it uses, and the methods it applies.
This led to the questions of how to effectively identify, acquire, represent, store, access, and use this design knowledge for generating explanations in interactive agents \cite{Goel1996a,Goel1996b}.
One solution lies in viewing the AI agent as an abstract device, equipping it with meta-knowledge about its design, and enabling it to introspect and generate explanations based on its understanding of its structure, behaviors and functions \cite{goel2011meta}.




There has been ongoing research into an Interactive AI agent's ability to provide self-explanation ~\cite{gunning2019darpa, tulli2024explainable}. In prior work on the Skillsync project for skill-based linking employers and colleges preparing prospective employees \cite{robson2022intelligent}, we used a Task-Method-Knowledge model of Skillsync to generate explanations of its reasoning and recommendations \cite{Goel2022}. A Task--Method--Knowledge (TMK) model captures an agent's design, knowledge, and reasoning processes into a unified structured representation \cite{rugaber2013gaia, murdock2008meta}.





With the rise of Large Language Models (LLMs) \cite{wei2022emergent}, Generative AI  methods have been integrated to enhance self-explanation in Interactive AI agents. In previous work on the SAMI project on connecting online learners with one another \cite{Goel2022,kakar2024sami,wang2020jill}, we integrated cognitive AI methods based on the TMK model of SAMI with generative AI methods to generate explanations of SAMI's reasoning and recommendations \cite{basappa2024social}.  




While these bodies of work serve as the background and context for our work, in the next section we describe how our work makes a novel contribution to the literature through generation of self-explanations for VERA, an interactive agent that supports inquiry-based modeling in the domain of ecology. In Section \ref{sec:method}, we first describe the TMK model of VERA as an interactive agent.  
We then combine this with generative AI methods to explore how VERA can introspect on its TMK self-model to provide reasoned explanations to a user’s query about VERA’s functioning.\\

\section{Methodology} \label{sec:method}

\begin{figure*}[h]
  \centering
  \includegraphics[width=\linewidth]{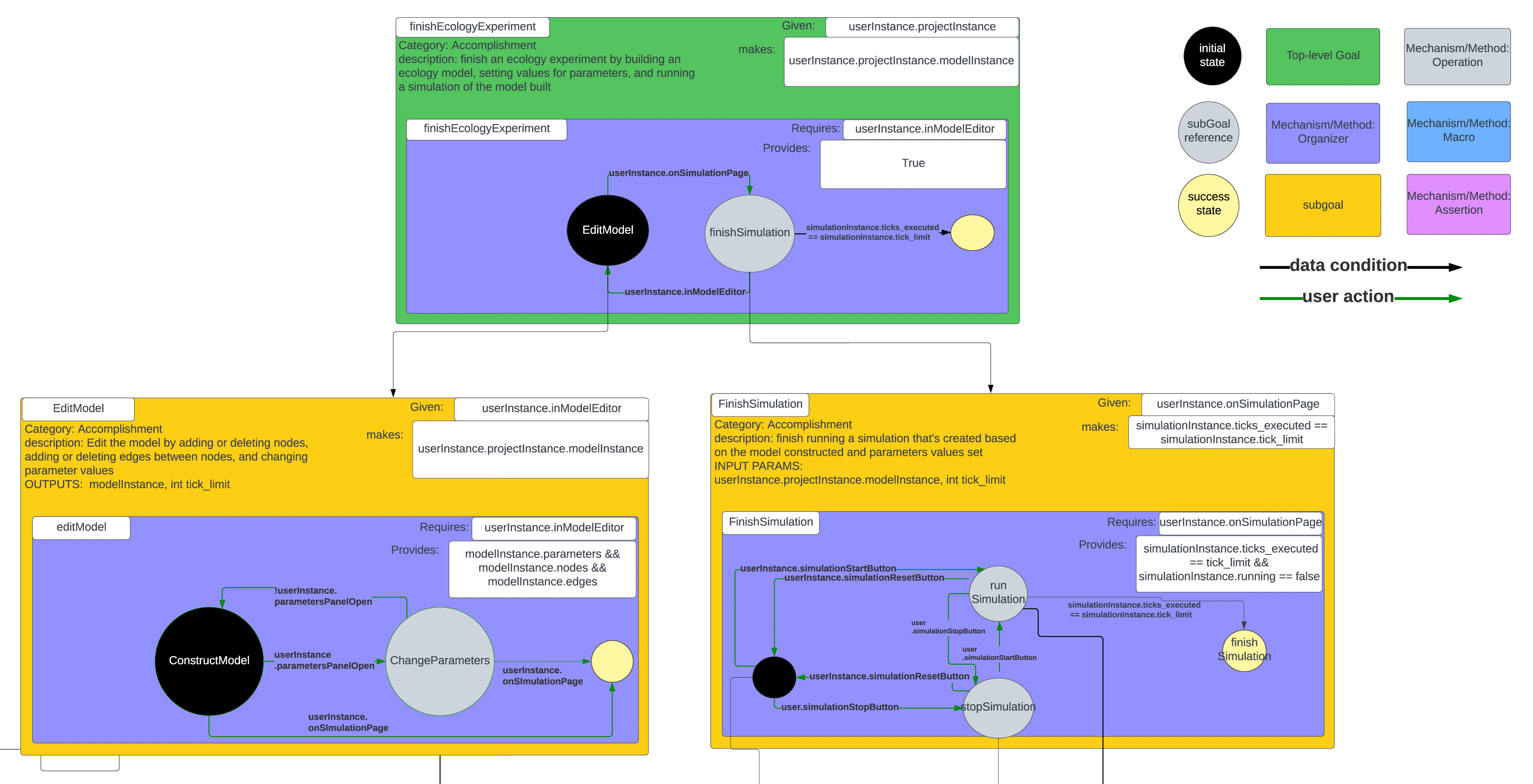}
  \caption{A portion of the TMK model of VERA, an interactive learning environment that supports inquiry-based learning in the domain of Ecology.}
  \label{fig:vera-tmk}
\end{figure*}

We present a novel approach to self-explanation in interactive agents such as VERA grounded in the agent's theory of its own mind. A theory of mind refers to an agent's capacity to ascribe mental states to others as well as to oneself. Here mental states refer to goals, desires, knowledge, beliefs, thoughts, emotions, etc. Recently theory of mind has emerged as a theoretical lens to understanding and designing human-AI interaction \cite{wang2024theory}.



\subsection{Theoretical Foundations for \\ self-explanations using TMK}

We posit that if an interactive agent has theory of its own mind, then it can use the self-theory to explain its reasoning and how the reasoning led to specific decisions. We use Task-Method-Knowledge (TMK) models to capture elements of an interactive agent's theory of its mind. We view the AI agent as an abstract device.
This device comprises a design with well-defined functions, constituent components with their own functionalities, and causal mechanisms that orchestrate these component functions to achieve the overall agent's goals.
Here, hierarchy refers to the layered structure of the design, causality describes the cause--and--effect relationships between components and functions, and teleology signifies the inherent goal-oriented nature of the design, see Figure \ref{fig:vera-tmk}.
Notably, TMK offers a natural mapping between its functions and tasks, and between its methods and mechanisms, aligning seamlessly with the proposed view of an interactive agent as an abstract device.

\subsection{Research Questions and Hypotheses}
\label{sec:RQs_RHs}
Based on this theoretical foundation, we formulate the following research questions (RQ) and corresponding research hypotheses (RH):

\begin{description}
    \item[RQ1:] How may an IA introspect on its design and explain its functioning?
    \begin{description}
        \item[RH1:] By representing the design as a TMK model, the IA can introspect on its design and explain its own functioning.
    \end{description}
    \item[RQ2:] How may an IA reflect on its design and explain its results for a given input instance?
    \begin{description}
        \item[RH2:] By processing through the TMK model, the IA can construct a derivational knowledge trace for the given instance and then generate an explanation by reflecting on the trace.
    \end{description}
\end{description}

In the following two subsections, we provide insights to these RQs and RHs.
First, from a cognitive AI perspective, we describe our approach for representing the interactive agent's design.
Then, by leveraging methods from generative AI, we describe how an IA introspects over its design and produces explanations about its functioning.
The implementation of cognitive and generative AI methods for self-explanations in VERA led to the development of the self-explanation module in VERA which we call \textit{Ask-TMK in VERA}. 
For the remainder of this paper, we shall simply refer to it as ``Ask-TMK''.

\subsection{Cognitive AI: TMK model of VERA} \label{sec:TMKmodel}
Ask-TMK's cognitive AI capabilities leverage VERA's Task Method Knowledge (TMK) representation—a comprehensive self-model encompassing goals, internal processes, states, concepts, relationships, and transitions. This teleological structure empowers Ask-TMK to actively monitor VERA's current state, reason about goal achievement, and systematically pinpoint the methods and concepts essential for fulfilling objectives \cite{rugaber2013gaia}.

To provide Ask-TMK with a structured knowledge representation of VERA, we manually constructed a TMK model—an abstract description of VERA's design.``TMK'' is an acronym for ``Task--Method--Knowledge'', three core aspects of any TMK model.
They are as follows:
\begin{itemize}
    \item \textbf{Task}. This part of the TMK model refers to VERA's objectives,   describing its aim, purpose, or the task being modeled. Tasks are expressed through the inputs (``givens'') and the resultant outputs (``makes''). For instance, in Figure \ref{fig:vera-tmk}, we consider VERA's task of ``Finishing an Ecology Experiment''. As the input to this task, a VERA project must be created, and the subsequent output is a conceptual ecological model. TMK models are inherently hierarchical, meaning that top-level goals of VERA can be decomposed into subgoals.  As shown in Figure \ref{fig:vera-tmk}, VERA's top-level goal (highlighted in green) is to ``Finish an Ecology Experiment''. To accomplish this, depending on the context, there are two immediate subgoals (highlighted in yellow): ``Edit a (conceptual ecological) Model'' or ``Finish a Simulation''. For more details about how VERA works, see our previous work \cite{an2018vera,an2020vera}.
    
    \item \textbf{Method}. This module of the TMK model describes how VERA accomplishes its Task. Methods are normally described by deterministic finite state machines (FSM) which in turn are defined by a set of states and transitions, see Figure \ref{fig:vera-tmk} (highlighted in purple). Similar to tasks, methods are also hierarchical. Therefore, top-level methods can be broken down into submethods.

    \item \textbf{Knowledge}. This final module of the TMK model corresponds to the definitions of the concepts and logical expressions used to specify the Tasks and Methods. This includes normal first-order logic operations and relations to connect with user supplied values \cite{rugaber2013gaia, murdock2008meta}. 
\end{itemize}

Using VERA's software documentation, a TMK model was manually created by core developers.
The amount of effort required to produce a TMK model is dependent on the level of abstraction to model the interactive agent. 
Initially TMK models are designed using a symbolic representation (see \ref{fig:vera-tmk}) and subsequently manually converted to a JSON representation.
Subsequent explanation generation utilizes these pre-built modules, resulting in a fully automated workflow.
To further streamline this process and reduce upfront investment, we plan to explore utilizing off-the-shelf software solutions for automated TMK module generation in future iterations.

\begin{figure*}[h]
  \centering
  \includegraphics[width=\linewidth]{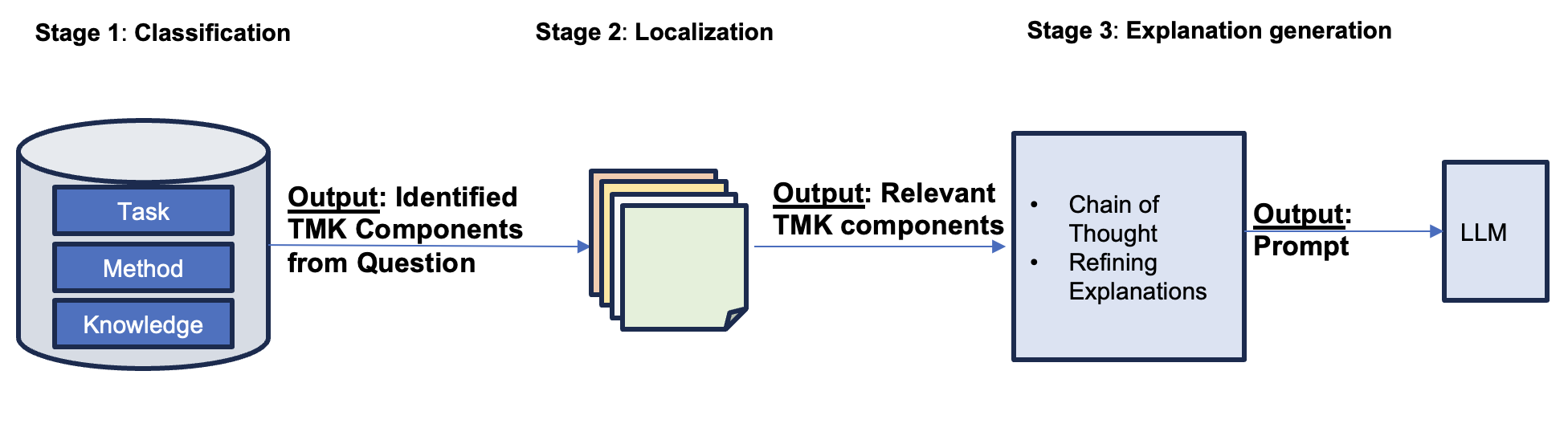}
  \caption{Combining Cognitive and Generative AI in VERA}
  \label{fig:convergence_primary_roles_v2}
\end{figure*}

\subsection{Generative AI for VERA self-explanations}
\label{sec:Generative_AI}

\subsubsection{ChatGPT, LangChain, and Chain-of-Thought}
We provide an overview of several Generative AI methods employed within Ask-TMK.
We focus on three key components: ChatGPT \cite{achiam2023gpt}, LangChain \cite{langchain}, and Chain-of-Thought\cite{wei2022chain}, highlighting their roles in generating user explanations based on VERA's TMK model.
We then go through a working example in Section \ref{sec:Combining_explanations}.

Ask-TMK leverages ChatGPT, specifically GPT-3.5 Turbo, to generate natural language explanations for users.
Upon receiving a user question, Ask-TMK utilizes the Large Language Model (LLM) to search and retrieve the relevant TMK documents.
Similar to prior work \cite{basappa2024social}, we use  
LangChain to create prompts that guide the LLM towards generating informative explanations.
Using the process of \textit{iterative refinement} \cite{selfrefine}, LangChain introspects over relevant documents from VERA's TMK model to answer user queries.

Ask-TMK leverages Chain-of-Thought to generate explanations with reasoning, for ``methods'' specific questions.
Chain-of-Thought is a reasoning technique that enables the LLM to explicitly reveal the steps it undergoes when arriving at an answer \cite{wei2022chain}. 
Ask-TMK integrates Chain-of-Thought during the reasoning stage by employing LangChain to construct prompts that guide the LLM to break down complex methods within the TMK model into subtasks and submethods.

    
 

\subsubsection{Experimental Setup}
\label{sec:experimental_setup}
The experimental setup involved configuring the GPT-3.5 Turbo model to generate responses, with constraints to ensure deterministic output.
Specifically, the responses were limited to a maximum of 1920 tokens, the temperature was set to 0, and verbose mode was disabled.
For document retrieval, a FAISS-based search system \cite{douze2024faisslibrary} was employed, configured with a k-value of 4 to return the top four most relevant documents. 
Document embeddings were created using OpenAIEmbeddings, and the search space comprised documents categorized as Task, Method, or Knowledge.
The k-value \cite{douze2024faisslibrary} refers to the number of nearest neighbors considered in a k-nearest-neighbor search, which is a common operation in similarity search algorithms.
Memory augmentation was achieved by incorporating the ``software\_qa\_prompt'' to facilitate the recall of previously presented information.
Lastly, as input to Ask-TMK, the self-explanation module received a ``question'' variable as input to generate its responses.

\subsection{Combining Cognitive and Generative AI}

Inspired by prior work \cite{basappa2024social}, we have chosen to benchmark VERA's self-explanation system using a bank of 66 questions that aim to test our research questions and hypotheses in Section \ref{sec:RQs_RHs}
\subsubsection{How does combining Cognitive and Generative AI generate explanations?}
\label{sec:Combining_explanations}

We demonstrate how VERA's innovative self-explanation system integrates Cognitive AI with Generative AI to produce detailed explanations. Figure \ref{fig:convergence_primary_roles_v2} depicts the collaborative operation of these two fundamental AI paradigms within the system. Cognitive AI plays a pivotal role in the initial phases of query processing, facilitating the structured identification of the pertinent TMK modules by enabling a teleological structure and organization of VERA's self-model as briefly outlined in section \ref{sec:TMKmodel}

Generative AI takes on a prominent role during the subsequent stage of explanation generation. Here, it utilizes the retrieved TMK components, potentially refining them to better suit the user's query context. This refined knowledge is then employed by the system's Large Language Model (LLM) component to generate a coherent and contextually appropriate explanation tailored to the user's needs.

Thus, this combination ensures that explanations are both accurate and contextually relevant, enhancing the user's understanding of complex queries.
The detailed explanation of each stage and how they interact is in Section \ref{sec:working_example}.
Additionally, we go over a working example with a question taken from our bank.

\subsubsection{A Working Example}
\label{sec:working_example}
We walk through an example question here taken from our bank of 66 questions. Consider the following scenario: \\ \\
\textbf{User question:} ``How can I best utilise the output of the system in VERA?''
\begin{enumerate}
    \item \textbf{Stage 1: Question Classification}

This stage is responsible for categorizing user questions to determine their relevance to VERA's internal model (TMK) and allocate resources efficiently for response generation. It operates as follows:

\begin{itemize}
    \item \textbf{Input:} The user question serves as input to a classifier powered by LangChain. This classifier uses pre-defined classes (outlined below) to categorize questions and identify the most relevant parts of TMK for answering.
    
    \item \textbf{Classification Process:} The classifier, utilizing GPT-3.5 Turbo, distinguishes question types and retrieves relevant models and corresponding documents based on tasks, methods, or knowledge within TMK.
    
    \item \textbf{Class Utilization:}
    \begin{itemize}
        \item \textbf{Mmodel Class:} This class, used for `Method' related questions, employs Chain-of-Thought Prompting during later stages to fetch relevant tasks and corresponding methods. It focuses on presenting intermediate steps within TMK, making it suitable for `How' questions.
        
        \item \textbf{Multimodels Class:} Handling all other question types, this class retrieves all relevant TMK documents without utilizing Chain-of-Thought during later stages. It aims to provide comprehensive responses covering various aspects of TMK.
        
        \item \textbf{Cant\_answer Class:} Dedicated to cases where the system cannot answer a question, this class ensures efficient resource allocation by redirecting such queries appropriately.
    \end{itemize}
    
    \item Based on this classification, the system determines which information from the TMK to provide to the next stages. Further, by tailoring response generation based on the specific information needs of each question type, this approach optimizes resource utilization and enhances the relevance and accuracy of responses.
    \item \textbf{Output of this stage for our working example:}  In this case, it classifies the question as ``Multimodels" and loads all the parts of the TMK. If a question is classified a ``Mmodels", only Task and Method parts of the TMK model is loaded: 
    \begin{itemize}
        \item \textbf{Pre-defined Class identified} - ``Multimodels"
        \item Method names: Loads various methods such as ``create simulation'', ``run simulation'', etc.
        \item Task names: Loads tasks like ``finish ecology experiment'', ``create simulation'', etc.
        \item Knowledge names: Loads knowledge names such as ``Ecology Model'', ``VERA'', etc.
    \end{itemize}

\end{itemize}
    \item \textbf{Stage 2: Localization}
    \begin{itemize}
        \item \textbf{Input:} This stage receives the classified question and the complexity factor, `k'-value from the previous stage. Please see \ref{sec:experimental_setup} for more details on k-value.
        \item The complexity factor influences the level of detail required in the explanation and correlates with the number of documents to be searched during FAISS search. \cite{douze2024faisslibrary}
        
        \item In this stage, FAISS similarity search \cite{douze2024faisslibrary} \ref{sec:experimental_setup} is employed to pinpoint the most relevant elements within the TMK, and LangChain is subsequently utilized to construct a prompt incorporating this relevant TMK information.
        \item \textbf{Output of this stage for our working example:} The system identifies and retrieves the most relevant TMK components along with their similarity scores. In the context of Ask-TMK, a similarity score represents a numerical value between 0 and 1 that indicates how relevant a specific element from the TMK model is to the user's question. FAISS provides us this similarity score for all of the k-documents.
        \begin{itemize}
            \item Ask-TMK: Explanation module within VERA. \textbf{Similarity Score: 65.16\%} 
            \item Ecology Model: Digital representation of an ecological system. \textbf{Similarity Score: 65.04\%}
            \item What if Experiment: Enables users to predict outcomes of future changes to an ecological system. \textbf{Similarity Score: 64.32\%}
            \item User: A person interacting with VERA to explore ecological systems and run simulations. \textbf{Similarity Score: 63.24\%}
        \end{itemize}
    
    \end{itemize}
    \item \textbf{Stage 3: Explanation Generation}
    \begin{itemize}
        \item \textbf{Input:} This stage receives the identified relevant TMK elements from the localization stage.
        \item It leverages LangChain once more to prompt the LLM to construct the final explanation for the user. We leverage LangChain to create a prompt providing all the necessary details including TMK and the context of the question to generate a natural language explanation tailored to the user's question and the ‘k’ value (complexity factor).
        \item Refinement of explanations happens as a separate step and a final answer is produced.
        \item \textbf{Output: Intermediate Steps} The LLM generates an answer in multiple steps, progressively refining the explanation. The intermediate steps are below:
        \begin{itemize}
            \item Analyze the results of ecological simulations and draw conclusions based on parameter changes.
            \item Leverage the ecological model for visualization, simulation, and ``what if'' scenarios.
            \item Gain a deeper understanding of ecological system dynamics and make informed decisions based on the output.
            \item Utilize the ``what if'' experiment feature to predict future outcomes by modifying parameters.
        \end{itemize}
        \item \textbf{Output:Refinement and Final Answer} The explanation is refined into a single, cohesive answer. The direct answer from the system is given below.
        \begin{itemize}
            \item ``The output of the system in VERA can be best utilized by the user to analyze the results of ecological simulations and draw conclusions about the behavior of the ecological systems based on the parameter changes they have made. As a user, you can visually represent the relationships between organisms and their environment, simulate interactions and behaviors within the model, and explore ``what if'' scenarios by running simulations with different initial conditions or changing parameters. This allows you to gain a deeper understanding of the dynamics of ecological systems and make informed decisions based on the output of the system. Additionally, VERA's ``what if'' experiment feature enables users to predict the outcome of future changes to an ecological system by modifying parameters and observing the resulting simulation, providing valuable insights for decision-making.''
        \end{itemize}
        \item As outlined previously in section \ref{sec:Generative_AI}, VERA leverages Chain-of-Thought for more intricate explanations, particularly when methods are involved. Chain-of-Thought enables the LLM to break down complex methods within the TMK into smaller tasks and subtasks, providing a more detailed explanation. 
        
        \item An example of the prompt used is provided in Appendix \ref{appendix}. The prompt was refined iteratively using Langchain's functionality, specifically the "refine" parameter within the load\_qa\_chain\cite{langchain} function. 


    \end{itemize}
\end{enumerate}

\section{Preliminary Results} \label{sec:results}
\subsection{Evaluation of the self-explanation Method}

We evaluated the self-explanation system (e.g., Ask-TMK, hereafter referred to as the system) implemented within VERA to assess its ability to provide informative and relevant explanations to user queries. This evaluation focused on the system's capacity to explain its internal workings and functionalities.

\subsubsection{Question Set and  Adaptation to VERA}

A set of 66 high-level, non-context-dependent questions was derived from established Explainable AI (XAI) question banks \cite{Liao_2020,sipos2023} and used in our previous work.
These questions were then adapted to VERA's specific context to ensure their relevance to the system's functionalities and user interaction.
We used the same set of questions to benchmark how VERA did with regards to our previous work.
The initial pool of questions was taken from established question banks from relevant research papers, focusing on those aligned with our prior work \cite{basappa2024social}. Further, the categorization of questions into relevant groups and the definitions of those categories was taken directly from the existing literature and question bank classifications used in prior works, such as those by Liao et al. (2020) \cite{Liao_2020} and Sipos et al. \cite{sipos2023}(2023).
SAMI developers, then, collaboratively reviewed these questions to ensure their relevance to SAMI's functionalities and objectives. This iterative process involved either directly accepting relevant questions or modifying them to better align with SAMI's specific context. The focus on relevance resulted in a variation in the number of questions across different categories, reflecting the inherent differences in the types of explanations SAMI can generate compared to other AI systems. These questions from our prior work were then taken by the developer for Ask-TMK in VERA and adapted to VERA's specific context in order to benchmark the performance of self-explanation in VERA.

\subsubsection{Evaluation Methodology}
\label{Evaluationmethod}
The evaluation process involved the following steps:

\begin{enumerate}
    \item Question Selection and Adaptation: As mentioned previously, relevant questions were selected from XAI question banks and adapted to VERA's specific functionalities and user interaction. Additionally, questions addressing VERA-specific aspects were created.
    \item Explanation Generation: Each of the 66 adapted questions was presented to VERA's self-explanation method via a user interface and the generated explanations were documented. 
    \item Evaluation Methodology: To assess the effectiveness of VERA’s self-explanation method in conveying information within a learning environment, we employed three established metrics commonly used to evaluate generative and cognitive AI systems: Recall, Precision, and Accuracy \cite{manning2008introduction, sasaki2007fmeasure, mitchell1997machine}  (Please see Table \ref{tab:explanation_metrics} for a definition of these metrics and what those ratings mean).
    In this initial assessment, we focused on evaluating explanations from an AI research perspective, excluding user-specific metrics. To evaluate VERA's responses, the Ask-TMK developer independently assessed each explanation against pre-defined criteria established from an AI research perspective\cite{mitchell1997machine, sasaki2007fmeasure,manning2008introduction}. These criteria focused on aspects defined above and the justification regarding why a certain rating was chosen was documented.  Another research scientist reviewed some of these initial ratings and the justifications for any discrepancies in the ratings were documented.

 Our future work will involve user-centered studies to evaluate comprehensibility by diverse user groups and refine VERA's self-explanation method for optimal user experience. 

While evaluating VERA using the same set of 66 questions previously employed with SAMI \cite{basappa2024social} suggests promise for generalizability, we acknowledge the need for further investigation. Future work will involve deploying VERA in diverse classroom settings to gather real-world data and comprehensively assess its generalizability across various learning environments.

This focus on real-world deployment will also allow us to delve deeper into the equity and bias aspects of VERA's self-explanation approach (Ask-TMK).  We will explore potential biases within the training data and consider how to ensure fairness and inclusivity in VERA's explanations across diverse user groups.
\begin{table} 
    \centering
    \caption{Explanation Metrics and Their Ratings}
    \begin{tabular}{|p{2cm}|p{6cm}|}
        \hline
        \textbf{Metric} & \textbf{Rating Descriptions} \\ \hline
        
        Recall & 
        Measures proportion of relevant information retrieved by self-explanation compared to total available. \\
        & \textbf{High:} Captures most relevant information. \\
        & \textbf{Medium:} Some relevant information missing or unclear. \\
        & \textbf{Low:} Significant gaps or inaccuracies. \\ \hline
        
        Precision & 
        Evaluates proportion of information directly addressing user’s query. \\
        & \textbf{High:} Highly focused and relevant. \\
        & \textbf{Medium:} Some irrelevant or minor inaccuracies. \\
        & \textbf{Low:} Substantial off-topic content or inaccuracies. \\ \hline
        
        Accuracy & 
        Assesses factual correctness of presented information. \\
        & \textbf{High:} Mostly correct and verifiable. \\
        & \textbf{Medium:} Some errors or inconsistencies. \\
        & \textbf{Low:} Significant inaccuracies or factual errors. \\ \hline
    \end{tabular}
    \label{tab:explanation_metrics}
\end{table}
    
\end{enumerate}
\subsection{Summary and Analysis of results}

The results have been summarized in Table \ref{tab:explanation_metrics}. We examine the performance of the self-explanation system the interactive agent, VERA, based on a user evaluation summarized in Table 1. The evaluation involved 66 questions taken from previous work as outlined earlier and categorized based on the type of information they sought.

\subsubsection{Overall Performance}

The self-explanation method achieved high recall, precision, and accuracy across most question categories, indicating its effectiveness in retrieving relevant information and generating accurate explanations.  

\subsubsection{Category--wise breakdown}
\begin{enumerate}
    \item \textbf{Input Questions (4)}: These questions focused on the VERA's training data and achieved perfect scores across all metrics.
    \item \textbf{Output Questions (22)}: This category, inquiring about how to utilize the VERA's output, had a slight decrease in precision (one medium score) compared to other categories. This was due to an occasional explanation that was accurate but not maximally helpful for optimal output utilization.
    \item \textbf{``How'' (Global) Questions (17)}: These questions aimed at understanding the general workings of the system. The system performed very well here, achieving high scores across all metrics. 
    \item \textbf{``Why Not'' Question (1)}: This category, with only one question, showed perfect performance.
    \item \textbf{``Others'' Questions (10)}: These questions covered various topics unrelated to the core functionality. The system performed well here, with high scores across all metrics. 
    \item \textbf{``Others'' (Context) Questions (3)}: These context--related questions received perfect scores across all metrics.
    \item \textbf{VERA Specific Questions (9)}: These questions focused on understanding specific outputs from VERA simulations. Again, the system exhibited high performance here.
\end{enumerate}
\begin{table*}[htb]
    \caption{Results of categorising all 66 questions used to evaluate the self-explanation method, along with a representative question for each category, their adaptation, and corresponding recall, precision, and accuracy scores}
    \resizebox{\textwidth}{!}{%
    \begin{tabular}{p{2cm}p{2.5cm}p{4cm}p{4cm}lll}
        \hline
        \textbf{Category} & \textbf{ \# of Questions} & \textbf{Example Question} & \textbf{Actual Question Tested} & \textbf{Recall} & \textbf{Precision} & \textbf{Accuracy} \\ \hline
        Input & \centering 4 & What kind of data does the system learn from? & What kind of data does VERA learn from? &  
 High - 4 & High - 4 & High - 4 \\ \hline
        Output & \centering 22 & How can I best utilize the output of the system? & How can I best utilise the output of the system?\newline How can I best utilise VERA’s output?\newline How can I best utilize the simulation outputs? & High - 22 & \makecell[l]{High - 21\\ Medium - 1} & High - 22 \\ \hline
        How (global) & \centering 17 & Is [feature] used or not used for the predictions? & Is simulation parameter used or not used in a simulation?\newline Is simulation behavior processes such as consuming, producing used or not used in running simulations? & High - 17 & High - 17 & High - 17 \\ \hline
        Why not & \centering 1 & Why/how is this instance not predicted?  & Why does my simulation not give an expected outcome? & High - 1 & High - 1 & High - 1 \\ \hline
        Others & \centering 10 & What are the results of other people using the system? & What are the results of other people using the system?\newline Would I be affected if other students use or not use VERA?\newline How will I be affected if other students use or not use VERA? & High - 10 & High - 10 & High - 10 \\ \hline
        Others (context) & \centering 3 & Who is responsible for this system? & Who is responsible for this system? & High - 3 & High - 3 & High - 3 \\ \hline
        VERA specific question & \centering 9 & Why did my simulation give this particular output? & Why did my simulation give this particular output? & High - 9 & High - 9 & High - 9 \\ \hline
    \end{tabular}%
    }
\end{table*}

\FloatBarrier
\subsubsection{Potential Areas of Improvement}
Overall, the self-explanation method demonstrates promising performance across most question categories. High recall, precision, and accuracy indicate that the system effectively retrieves relevant information and provides accurate explanations. \\
As pointed out earlier in \ref{Evaluationmethod}, the current system has undergone preliminary evaluation led by the developers, focusing on AI research perspectives. It has not yet been deployed in classroom environments. We acknowledge the potential for unintentional biases stemming from our deep familiarity with the Ask-TMK system's internal mechanisms, which may have influenced question framing and answer interpretation.  It is anticipated that deployment in real classrooms will introduce a layer of human-centric evaluation currently lacking, potentially yielding divergent insights. Future research will prioritize the incorporation of these critical human evaluations to improve the system's relevance and performance within educational settings. \\
For future work, we plan to: 
\begin{enumerate}
    \item Test the system with more questions to determine if precision scores vary or if we encountered an occasional outlier.
    \item Conduct user studies to understand how the self-explanation system performs with different user groups.
\end{enumerate}

\section{Conclusion}
 The Ask-TMK module in VERA uses a theory of VERA's mind to explain how it works through question answering. Ask-TMK's theory of VERA's mind is captured in the language of Task-Method-Knowledge (TMK) models that specify how VERA uses its domain knowledge and reasoning methods to achieve its goals.  We tested the Ask-TMK self-explanation system within VERA with the question bank established in previous work.  Our preliminary analysis shows that the self-explanation system effectively leverages cognitive AI's structured knowledge for information retrieval and generative AI's capabilities to deliver relevant and accurate explanations. The system maps user queries to the relevant Task, Method, and Knowledge components within the TMK model, thereby generating responses that explain how VERA works. In our use case, this integration enables factually accurate, complete, and precise explanations and demonstrates promising performance across various question types.
\section{Acknowledgments}

We are grateful to Dr. Spencer Rugaber at Georgia Tech's Design Intelligence Laboratory for his invaluable insights into TMK models and modeling. This research has been supported by NSF Grants \#2112532 and \#2247790 awarded to the National AI Institute for Adult Learning and Online Education.

\FloatBarrier

\bibliography{Background}  

\appendix

\section{Appendix}
\label{appendix}

\subsubsection*{Prompt for Multi-model class}
\begin{lstlisting}
multi_models_desc = """multimodels: multimodels questions involve your system's knowledge, concepts, tasks, and methods. Your system has the following concepts in a JSON file: {Knowledge_names}. Your system performs the following tasks in a JSON file: {Task_names}. Your system has the following methods in a JSON file: {Method_names}.The Templates for example `multimodels' questions might be `Why do you need [concept]?' or 'What do you do with [concept]?' or 'How do you do with [concept]?"""
\end{lstlisting}

\subsubsection*{Multi-Model Answer Prompt}
\begin{lstlisting}
multi_models_answer_prompt = PromptTemplate(input_variables=[`software_qa_prompt', `context_str', `question'], template="""{software_qa_prompt}. The JSON or XML given below contains information about the concepts, objects and their properties you track in your system, the tasks you perform and their parameters, and/or methods you use to perform tasks.{context_str} The user asks the following question: `{question}'. Please follow these precise guidelines when proving a response.
**Answer the user's question based on the above JSON files only, please forget what user has asked earlier. Please treat each {question} as completely new and completely unrelated to any previously asked question.Please answer the question in a concise and informative way, in a human-friendly natural language format, aiming for 1-2 sentences. Please avoid technical terms such as "process tick", "execute tick" and make it simple for any AI researcher to understand using simple words and sentences. If you need more information to provide a 
complete answer, you can indicate that to the user. Your goal is to be user-friendly. Try to answer each {question} from a fresh perspective assuming the user has no knowledge of what they are asking 
even if they have asked the question earlier. However, please stay to the point and concise while answering. If the existing answer cannot be refined further, state the final answer without refining further. Focus on providing an accurate answer that directly addresses the user's 
question. Do not including irrelevant information that do not relate to the question. If the answer is long, please paraphrase and summarize in 1-2 short sentences only offering user more details if they request it. If you cannot find information in any of the JSON files, please avoid making up answer and say you do not know. Ask the user to ask questions related to functionality of Vera only.**
""")
\end{lstlisting}
\balancecolumns
\end{sloppypar}
\end{document}